# Language-model groups overstate consensus when replaying human deliberation on a reasoning task

**Tengfei Shao***

Global Education Center, Waseda University, Tokyo, Japan

* Correspondence: tengfei.shao@toki.waseda.jp

## Abstract

Full-consensus rates are often treated as indicators of collective cognition, yet depend on how participation and final states are operationalized. We replayed 100 held-out human Wason groups with matched large language model (LLM) agent groups, seeding one belief-anchored agent per participant's pre-discussion answer and scoring agents and people with the same code. Across human scoring definitions, estimates ranged from 24.0% to 57.0%; about one fifth of participants never posted, whereas agents almost always did. Agent groups remained more consensual in two post-unblinding sensitivity analyses: the submit-based comparison (n = 98) yielded gaps of 34.0 and 43.9 percentage points for chat and reasoning modes, and the participation-matched comparison (n = 45) yielded gaps of 34.1 and 44.4 points. These complementary routes reduced different measurement asymmetries yet converged within 0.5 percentage points. The gap persisted without early stopping and under a reparameterization removing the memorizable answer; reasoning-mode groups then agreed nearly unanimously, mostly on incorrect answers. Simulated consensus did not track collective accuracy, and belief-anchored agent groups were biased estimators of the human group-outcome distribution in this setting. These analyses provide a scoring-explicit basis for assessing simulated-group estimates of human deliberative outcomes.

## Introduction

Deliberation is not simply a route to unanimity. Human groups can pool information, correct individual errors, and improve on their starting positions, but they can also preserve minority views, partial commitments, and unresolved disagreement. How often they end in full agreement is therefore a group-level estimator rather than a self-defining property of deliberation. Its value depends on which participants count as active and how unsubmitted or unchanged final states are represented, so operational choices can alter the apparent prevalence of agreement and the inferences drawn from it about collective cognition. In the DeliData Wason groups analyzed here, full consensus was 24.0% under corpus carry-forward scoring (n = 100), 52.0% under submit-based scoring (n = 98), 57.0% under active-only scoring (n = 100), and 51.1% among lurker-free groups (n = 45). About one fifth of participants never posted, and under an all-members criterion a group with a silent member whose carried-forward state differs from the others is non-consensual by definition. The Wason card selection task has a logically correct solution, although its difficulty and sensitivity to framing are long documented (Wason, 1968; Cheng & Holyoak, 1985; Oaksford & Chater, 1994). This scoring sensitivity makes the operationalization of consensus part of the validity argument, consistent with the construct-validity tradition and with guidance on measurement practice (Cronbach & Meehl, 1955; Flake & Fried, 2020).

This problem becomes consequential when large language model (LLM) agents are used as synthetic participants. For psychologists who report or review agent-group outputs as evidence about human deliberation, the decision at stake is whether the scoring rule and the participation structure warrant reading simulated group outcomes as evidence about people. We describe human-group behaviour in this Wason setting, not a universal law, as the target our agents must reproduce, and the signature we test combines whether groups improve, whether they reach full consensus, and whether an agreed answer is right or wrong.

This evidential use is increasingly common, which makes the signature newly important for psychology. Synthetic-participant and silicon-sample approaches ask whether language models, conditioned on demographic, attitudinal, or role information, can stand in for people (Park et al., 2023; Horton, 2023), and can approximate some human response distributions and replicate selected experimental patterns (Argyle et al., 2023; Aher et al., 2023). Recent validations sharpen the caution: model samples can match group means while collapsing the response variance and entropy that make a distribution human (Bisbee et al., 2024; Dominguez-Olmedo et al., 2024), persona conditioning flattens within-group diversity and can misportray whole identity groups (Wang et al., 2025; Santurkar et al., 2023), and models do not reliably track human response biases (Tjuatja et al., 2024). Whether model outputs are evidence about human minds remains contested (Dillion et al., 2023; Grossmann et al., 2023). The core issue is not whether LLMs produce human-like text but whether they preserve the psychological structure of the behavior being studied.

Reproducing a group's collective cognition is a harder test than reproducing individual responses. Collective intelligence in the sense of Woolley et al. (2010) is a latent factor that predicts a group's performance across many tasks; what concerns us is narrower, the trajectory of a single deliberation, who changes, who resists, and whether the shared answer is correct. Collective accuracy is driven by preserved diversity and independence rather than by agreement: cognitive diversity can outperform individual ability (Hong & Page, 2004), social influence that narrows diversity degrades the wisdom of crowds (Lorenz et al., 2011), and the majority answer is often not the accurate one (Prelec et al., 2017). Agreement is therefore not the goal, and over-convergence is a failure mode rather than a success. This matters wherever agent groups model deliberation, since independent evidence already shows LLM agent groups drifting toward conformity and groupthink (Zhang et al., 2024), degenerating into premature convergence in debate (Liang et al., 2024), and developing group-level biases not reducible to any individual agent (Ashery et al., 2025). If simulated groups converge too readily, they replace a human pattern of partial agreement with a model pattern of excess unanimity.

We therefore ask a construct-validity question: do belief-anchored LLM agent groups preserve the consensus structure of the human groups they are meant to stand in for? The Wason card selection task is a useful test bed because it pairs a real human discussion corpus with a ground-truthed answer. People often begin with wrong or partial beliefs, and discussion can improve, preserve, or worsen them. Ground truth does not make convergence detectable, since convergence is defined on the final answers alone; what it adds is the ability to decompose a consensus into correct versus wrong, distinguishing agreement on the solution from agreement on an error. Prior multi-agent deliberation work either compares agents on subjective questions where that decomposition cannot be defined (Chuang et al., 2025) or matches agents to real participants on a collective task without simulating the dialogue itself (Qian et al., 2025).

We preregistered a held-out replay of 100 DeliData Wason groups (Karadzhov, Stafford and Vlachos, 2023) after 400 calibration groups (osf.io/5jp7s), instantiated one belief-anchored agent per real participant from that participant's pre-discussion answer, crossed three role-fidelity scaffolds with three seeds and two inference modes of the same pinned deepseek-v4-flash served build, and scored human and simulated groups with the same code. The belief anchor was essential: unconstrained models can solve the puzzle far above the human individual baseline, which would make any group gain a model-competence artifact, so the design instead asks whether agents can deliberate from the beliefs real participants brought into discussion.

The answer is cautionary. Agent groups over-converged on full consensus far above the human groups they replayed; much of the raw gap reflects human non-participation and the carry-forward measure, but a substantial paired over-convergence remained on the participation-matched lurker-free groups for both inference modes. The stopping rule and recognition of the canonical answer did not explain this pattern. Over-convergence persisted with early stopping disabled and under an isomorphic reparameterization, where the reasoning mode still reached near-total agreement, predominantly on incorrect card sets. Elevated consensus also appeared in a cross-family open-weights model. Role fidelity reduced it relative to no scaffold, but per-turn belief memory did not. This preregistered, ground-truthed test is bounded to one served model in chat and reasoning inference modes, one cross-family robustness model, and one objective reasoning task, and its implication is a construct and protocol validity limit: high simulated consensus is not a reliable indicator of collective accuracy, so consensus, opinion spread, or minority survival read off these agent groups can misstate what a human group would show. We do not claim a general human mechanism; we show that in this protocol the simulated consensus endpoint over-states agreement.

## Results

### Agents over-state group improvement modestly and full consensus severely

The held-out confirmatory run replayed 100 DeliData Wason groups (Karadzhov, Stafford and Vlachos, 2023) across three role-fidelity scaffolds, three seeds, and two inference modes, deepseek-chat and deepseek-reasoner (1,800 cells). Both API aliases returned the same served model identifier (deepseek-v4-flash) and system fingerprint, identifying two inference settings of one served build (Methods). Figure 1 summarizes three stages: the human benchmark and sealed split, controlled belief-anchored replay, and same-code scoring. The comparison was preregistered (osf.io/5jp7s), and an integrity audit found zero leakage between held-out and calibration splits (Supplementary Methods). Seven deepseek-reasoner cells (0.4%) failed and were excluded rather than backfilled, leaving 100 groups per model-by-arm combination. The preregistered primary arm is fidelity_memory. Same-code human signatures on the same 100 groups were S1 improve 41.0%, S2 worse 4.0%, and S3 full consensus 24.0% under corpus carry-forward; across-arm S1–S3 rates appear in Supplementary Table S4. Because identical code does not ensure identical measurement, we also report participation-restricted human consensus rates. Native DeliData rates (S1 63.6%, S2 worse 17.6%, S3 24.0%) differ from same-code human rates on S1 and S2 because our analysis uses exact-set-match team performance; S3 is accuracy-independent and matches. Statistics below are group-clustered and Holm-corrected across the confirmatory family.

Language-model groups overstate consensus when replaying human deliberation on a reasoning task

*belief-anchored LLM agents over-converge on full consensus vs the human groups they replay · red = human, teal = LLM agents*

**Figure 1. Study overview: belief-anchored agent groups over-converge on consensus.** The numbered stages show the human benchmark, belief-anchored replay, and score-and-compare core. Stage 1 starts from 500 DeliData groups (1,974 participants and 14,003 utterances) solving the Wason four-card task. A fixed hash assigns 400 groups to calibration, which tunes only S1 group improvement, and seals 100 preregistered groups for held-out evaluation. Belief anchoring is used because an unconstrained model can solve the task much more accurately than humans: each agent is therefore seeded from one participant's pre-discussion answer so that the replay tests deliberation from the beliefs people actually brought to the group. Stage 2 instantiates one agent per participant while withholding the answer key, human discussion, post-discussion answers, and peers' private pre-answers. Agents speak or pass each round under three scaffolds, three seeds, and two inference modes, yielding 1,800 confirmatory cells; Qwen3-14B is an additional cross-family check. Agent lurking was 0.2% compared with 20.1% among humans. In the primary arm, chat and reasoner groups took 4.4 and 2.3 rounds, respectively, to reach a shared belief, and retained the injected belief in 34.6% and 17.1% of final answers. Stage 3 scores human groups and agent replays with the same code on S1–S4 and pairs them by group. Stage 3 (c.1) shows that agent groups reach near-total full consensus where human groups reach only partial consensus; the participation-matched result (in the 45 groups in which every human participant spoke, full consensus was 51.1% for humans, 85.2% for chat, and 95.6% for reasoner, paired gaps of +34.1 and +44.4 percentage points) and the all-groups cross-family Qwen3-14B check (65.7% against the 24.0% human carry-forward rate) are shown here in rounded form, with exact estimates and inferential detail in Figure 2 and the text. Full consensus records identical final answers, not correctness: on the isomorphic task the reasoner reached near-total consensus but 74.0% of all groups agreed on a wrong answer. The over-convergence persisted when early stopping was disabled, under isomorphic reparameterization, and in the cross-family check; the peer-hidden control changed chat consensus from 31.0% to 87.0% when peer messages were visible, compared with 91.0% to 100.0% for reasoner. Under the preregistered compatibility rule, in which a signature counts as reproduced only when the human value lies within the simulated 95% group-bootstrap interval, neither signature is reproduced within interval: agents over-state group improvement modestly and full consensus severely, illustrating a construct-validity limit: high simulated consensus does not by itself establish human-like collective accuracy.

Agents over-stated both signatures, modestly for group improvement and severely for consensus. Group improvement is the incidence of positive mean-member accuracy change, not the classical assembly bonus of beating the best member (Steiner, 1972; Laughlin et al., 2006). H1 tested whether simulated incidence exceeded a preregistered 0.5, a fixed decision threshold rather than a chance model. Chat agents improved in 60.7% of groups and reasoner agents in 85.3% (Fig. 2a), both exceeding 0.5 (chat +0.107, p_holm = 0.0029; reasoner +0.353, p_holm < 0.001), supporting H1 for both modes (Table 1). Under the preregistered compatibility rule, reproduction requires the human rate to fall within the simulated 95% group-bootstrap interval. Neither S1 nor S3 qualified: the human S1 rate of 41.0% lies below both intervals (Supplementary Table S2), so improvement was also over-stated, though less than consensus. A non-registered same-code contrast is metric-fragile: human incidence rises from 41.0% with exact-set scoring to 77.0% with graded Jaccard, which credits partial overlap with the correct two-card set unlike all-or-none exact-set scoring. We therefore do not claim that agents deliberate better than the humans they replay.

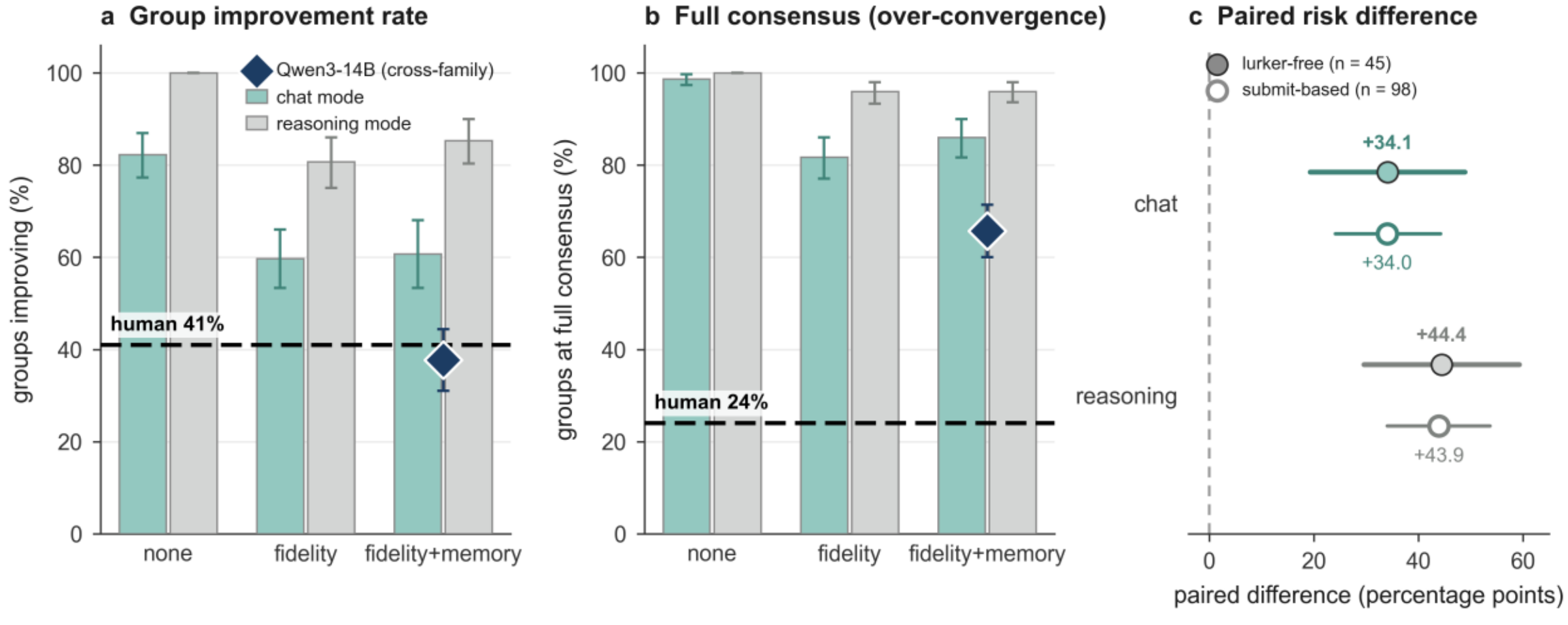

**Figure 2. Belief-anchored agent groups inflate positive group accuracy change but over-converge on consensus.** Group improvement rate (a) and full-consensus rate (b) on the 100 held-out groups, by scaffold arm and inference mode (chat, reasoning, both served from the deepseek-v4-flash build), with the same-code human carry-forward rate as a dashed reference; error bars in (a) and (b) are 95% group-bootstrap confidence intervals, and participation-restricted human full consensus is 52 to 57%. Agents show positive accuracy change above the human 41% in every arm, and reach full consensus far above the human 24% carry-forward rate. The diamond marks a cross-family robustness check on a different open-weights family (Qwen3-14B, not an independent-sample replication) on the primary arm, which shows the same over-convergence pattern (b) without inflating the accuracy-change incidence (a). (c) The headline participation-matched paired over-convergence: the per-group risk difference of agent minus human full consensus with 95% group-bootstrap intervals, filled markers for the lurker-free comparison (n = 45) and open markers for the submit-based sensitivity

comparison (n = 98); both sensitivity estimates are close and exclude zero for chat (+34.1 and +34.0 points) and reasoning (+44.4 and +43.9 points).

## Agents over-converge on full consensus

Full consensus records whether every final answer is identical, not whether the shared answer is correct, so it is independent of the accuracy metric. Human and agent measurements nevertheless differ. DeliData's carry-forward tracker holds a disengaged participant's last selection, whereas every agent is force-elicited; identical code is not identical measurement. Human full consensus is 24.0% under carry-forward, 57.0% among participants who posted at least one message, and 52.0% from actual submitted answers (n = 98). About 20% of participants never posted, so one silent member with a stale heterogeneous state mechanically breaks small-group unanimity. Among the 45 groups in which everyone spoke, participation-matched human full consensus is 51.1%.

The headline effect is the paired risk difference in lurker-free groups. Chat agents reached 85.2% full consensus against the human 51.1%, a +34.1 percentage-point difference (95% CI [19.3, 48.9]); reasoner agents reached 95.6%, a +44.4-point difference ([29.6, 59.3]). Both intervals clearly exclude zero. A complementary submit-based sensitivity paired each group's seed-averaged agent outcome with actual human submissions among groups with at least two submitters (n = 98). Excesses were +34.0 percentage points (95% CI [24.1, 44.2]) for chat and +43.9 points ([34.0, 53.7]) for reasoner, preserving over-convergence without the carry-forward tracker. The larger unadjusted all-groups gaps, +62.0 points [53.0, 70.7] for chat and +72.0 points [63.3, 80.3] for reasoner, both $p_{holm} < 0.001$, mix over-convergence with participation and carry-forward differences. The preregistered confirmatory H2a test is the all-groups paired comparison, which supports H2a; the lurker-free and submit-based values are participation-matched and measurement-based sensitivity estimates adopted during revision after unblinding, and we headline them because they are the more interpretable quantities (Supplementary Methods). Cohen's h (chat 1.35, reasoner 1.71) is descriptive and ignores pairing. Small-group selection does not explain the gap because even fully participating human groups reached consensus only about half the time.

## Over-convergence is not explained by stopping rules or task memorization

Because simulations halt when consensus settles or every agent passes, unlike human discussions, stopping could inflate consensus. We therefore disabled consensus and all-pass early stops in the primary arm, ran every group to the fixed horizon, and retained final elicitation. Full consensus was 87.0% for chat and 100.0% for reasoner (n = 100 groups each), essentially unchanged from 86.0% and 96.0% and far above every human measure (Fig. 3). A weaker selected-subset floor among cells not stopped on consensus remained 75.0% for chat and 94.3% for reasoner (Supplementary Table S2). Concordance therefore did not depend on implemented early-stopping rules. The fixed horizon and forced final elicitation remain, so this bounds rather than removes protocol influence; results are also stable across seeds and after excluding truncated cells (Supplementary Methods and Supplementary Table S2).

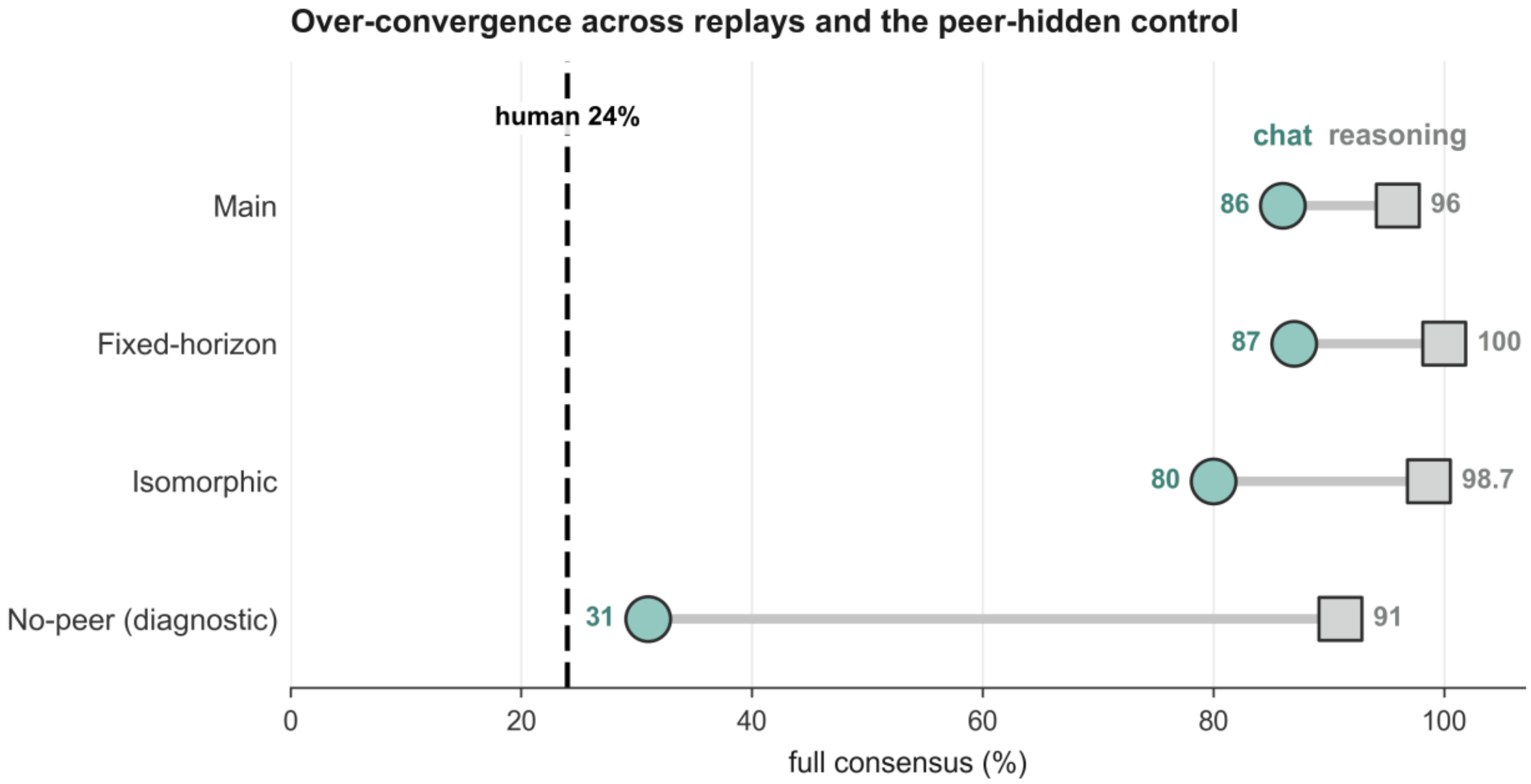

**Figure 3. Over-convergence persists across replays and separates from peer interaction.** Primary-arm full-consensus rate for the chat (circle) and reasoning (square) modes under the main run, the fixed-horizon replay (early stops disabled), and the isomorphic reparameterization, all far above the human 24% carry-forward reference (dashed). The no-peer solitary run is a diagnostic control, not a robustness check: hiding peer messages collapses the chat mode toward the human rate while the reasoning mode stays high, so the chat mode is strongly sensitive to peer visibility in this classic-task control. Exact rates (n = 100 per cell) are in Supplementary Table S5; the subjective-task under-convergence contrast is reported in Supplementary Note 1.

A second objection is recognition of the memorizable classic Wason answer. The preregistration (Section 4.5) therefore specified a held-out isomorphic reparameterization: each group's four cards were remapped to neutral tokens (maple, birch, lantern, candle) while preserving logical structure, removing the canonical letter-and-number answer. Full consensus persisted at 80.0% for chat and 98.7% for reasoner (n = 100 groups each), far above the human rate, and concordance did not track correctness. For the reasoning mode, correct consensus fell from 84.0% of groups on the classic task to 24.7% while wrong consensus rose from 12.0% to 74.0%, a paired within-group increase in wrong-consensus incidence of +62.0 percentage points (95% group-bootstrap CI [+53.3, +70.7], exact sign-flip permutation $p < 0.001$, n = 100 paired groups), so the agents still reached near-total agreement but predominantly on incorrect card sets; for chat, correct consensus was 36.7% and wrong consensus 43.3%. The reasoning mode's classic-task accuracy is therefore consistent with imported task competence or familiarity, not proven memorization.

This test is bounded because the neutral tokens are ordinary English words and the reparameterized rule is less natural. It establishes robustness to surface reparameterization, not memorization in isolation or a pure correctness-independent measurement: over-convergence persists without the memorizable answer and does not track correctness. Token semantics may shape the shared answer, but full consensus remained high across both task surfaces (chat 86.0% to 80.0%, reasoner 96.0% to 98.7%) and above the human rate under every measurement definition.

A cross-family robustness check replayed the same 100 held-out groups with locally served open-weights Qwen3-14B in the primary arm across three seeds. This is not an independent-sample replication because it reuses the same groups and one arm; its p-value is nominal and exploratory. Qwen full consensus was 65.7% against the human 24.0% carry-forward rate, a +41.7-point paired excess ($p < 0.001$, nominal). This all-groups carry-forward estimate is comparable to the deepseek gaps of +62.0 and +72.0, not the participation-matched lurker-free values. Qwen did not inflate improvement incidence (37.7% versus human 41.0%), making that result model-dependent, whereas over-convergence extended to this second open-weights family. Across all groups, Qwen reached 42.0% wrong consensus and 23.7% correct consensus, over-agreeing on errors and solutions.

## Consensus can be correct or wrong, and peer visibility acts differently by mode

Full consensus can land on a right or a wrong answer, and we report the split as a share of all groups so the two shares sum to the full-consensus rate. Of all human groups, 11.0% agreed on the correct card set and 13.0% on a wrong one (45.8% and 54.2% among the groups that reached consensus). Simulated groups over-produced both kinds: of all chat groups, 54.0% reached correct consensus and 32.0% wrong consensus, and of all reasoner groups on the classic task, 84.0% reached correct consensus and 12.0% wrong consensus (Fig. 4c). The chat mode over-agrees, including on errors, whereas the reasoning mode on the classic task concentrates on the correct answer, an accuracy that the isomorphic result shows is not intrinsic to the over-convergence. A reader treating either simulation as a human proxy would draw opposite conclusions about how often groups agree on something false.

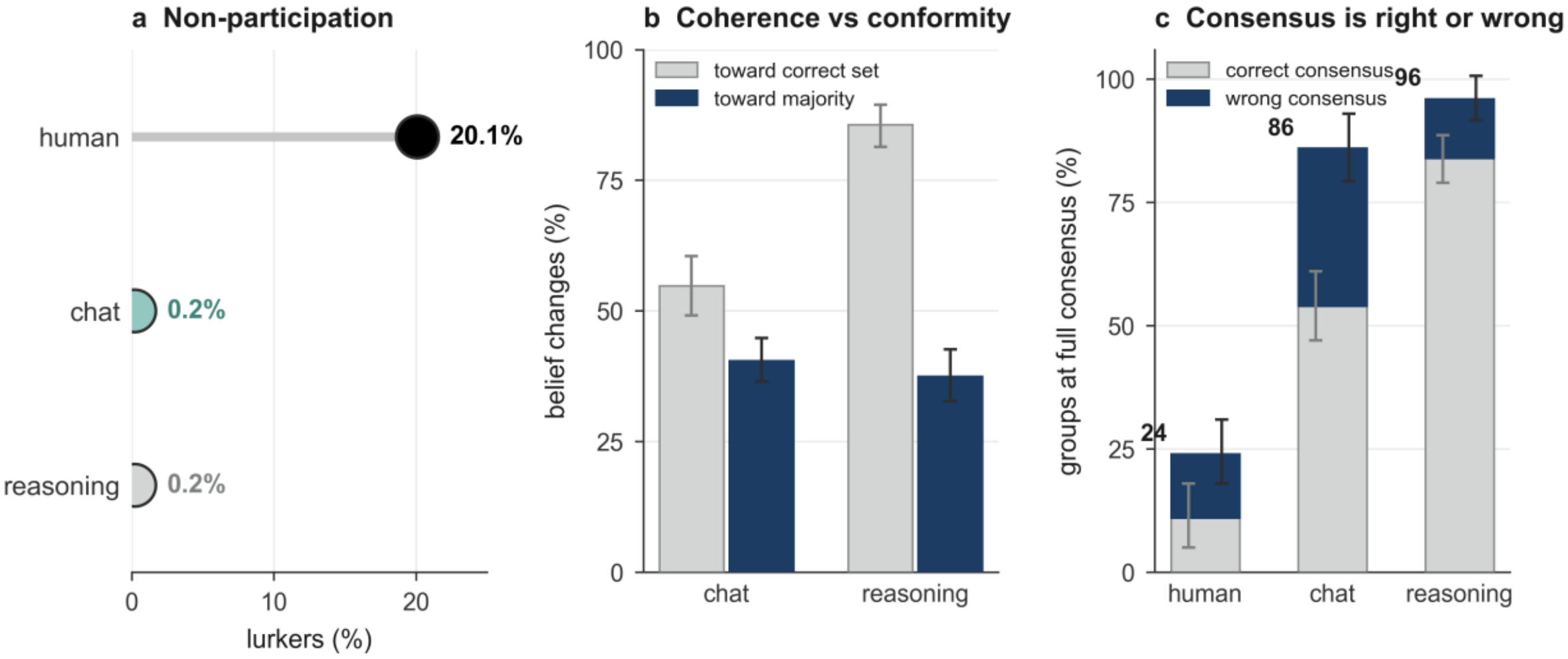

**Figure 4. Process and outcome signatures accompanying the over-convergence.** (a) Simulated agents almost never lurk, unlike the roughly 20% of human participants who never speak. (b) Reasoning belief changes move predominantly toward the correct card set, whereas chat changes mix that with movement toward the group majority; the two targets overlap when the majority already holds the correct set, so they are reported as correlational process signatures, not a demonstrated cascade. (c) Correct and wrong full-consensus shares stacked to the total full-consensus rate (of all groups): the bar height is the consensus rate and its split shows how much of that agreement is on wrong versus correct answers. Chat agrees on wrong answers more often than humans, while reasoning on the classic task concentrates on the correct answer; under the isomorphic task its consensus moves onto wrong sets (see text). Error bars in (b) and (c) are 95% group-bootstrap confidence intervals. Persona-retention rates are reported in the text and Supplementary Table S2.

Within the same served build, the reasoning inference mode reached full consensus more than the chat mode across every comparison: the primary arm, the no-scaffold arm, and the isomorphic task. We report this as a descriptive contrast between two inference settings of one build, not a capability gradient, because the two endpoints differ in decoding and served configuration and we have no independent capability manipulation. On the isomorphic task the reasoning mode's agreement lands predominantly on wrong answers, so its higher concordance there is not an accuracy advantage.

To test how much of that agreement depended on peer visibility, we compared a peer-hidden, fixed-horizon replay with its matched peer-visible condition on the classic task, all other settings held constant, so the peer-hidden condition measures agreement among agents separately re-solving the task rather than agreement produced through exchange. For deepseek-chat, full consensus was 31.0% without peer messages (26.3% correct, 4.7% wrong), below the human lurker-free rate of 51.1%, against 87.0% with peers, where wrong consensus rose from 4.7% to 32.0%. For the reasoning mode it was already 91.0% without peer messages, all correct, against 100.0% with peers (88.3% correct, 11.7% wrong). The two modes therefore differ in sensitivity to peer visibility (Fig. 3): holding the fixed horizon constant, the peer-visibility effect on full consensus is +56.0 points for chat (31.0% to 87.0%) against +9.0 points for reasoning (91.0% to 100.0%), a mode-by-visibility difference-in-differences of +47.0 percentage points (95% group-bootstrap CI [+34.7, +58.7], exact sign-flip permutation $p < 0.001$, $n = 100$ paired groups), indicating a substantial change for chat but little change for reasoning, with some reasoning-mode consensus instead shifting from correct to wrong. Because this control was run only on the classic task, and because high peer-hidden agreement in reasoning may reflect task-specific competence, it does not establish a general interaction mechanism or attribute the isomorphic-task wrong consensus to peer interaction.

## Role scaffolding bounds the mismatch, and agents almost never stay silent

H3 asked whether the scaffolds monotonically reduce distance to the human signatures, with the predicted order none > fidelity > fidelity_memory. Page's trend test is significant in both inference modes (chat $L = 1267.5$, $p < 0.001$; reasoner $L = 1250.5$, $p < 0.001$), reflecting the large gap between no scaffold and either scaffolded arm, but the predicted strict order does not hold. On the registered S1–S4 signature distance to the human vector, plain fidelity, not the calibrated fidelity_memory arm, sits closest to human behaviour in both modes, an ordering robust to computing distance on S1–S3 or S1–S4. On full consensus by arm (Fig. 2b), adding a role-fidelity instruction lowers it (chat 98.7% to 81.7%, reasoner 100.0% to 96.0%), whereas adding per-turn belief re-injection on top leaves it equal or slightly higher and beats plain fidelity in only 32 of 100 chat groups and 17 of 100 reasoner groups. Giving agents an explicit memory of their own starting position therefore did not add to the reduction in full consensus and instead slightly reversed it. We classify the monotonic-order hypothesis as unsupported.

We report process signatures from the transcripts that accompany the outcome as correlational rather than causal. Real DeliData groups contain about 20.1% lurkers, participants who never post, but simulated agents almost always speak: the lurk rate was 0.2% for both inference modes in the primary arm (Fig. 4a), so agents do not reproduce human non-participation, and every agent's belief is exposed to challenge on nearly every turn. Anchor retention differs sharply: only 17.1% of reasoner agents held their assigned belief for the whole discussion, against 34.6% for chat (Supplementary Table S2), so the reasoning mode abandons its injected persona far more readily, a gap that makes its classic-task accuracy hard to separate from imported competence (Discussion). Reasoner groups reached a single shared live belief at some round in 95.3% of groups against 62.0% for chat, and did so earlier: by round two, 76.0% of reasoner groups had already held a shared belief against 26.7% for chat, whose trajectory rises in a long tail (Supplementary Fig. S1). Decomposing belief changes by target, 85.6% of reasoner changes moved toward the logically correct card set and 37.5% toward the current majority; for chat the split was 54.8% and 40.5% (Fig. 4b). These categories overlap when the majority already holds the correct set. The pattern is consistent with the reasoning mode resolving disagreement toward the correct answer and the chat mode mixing this with movement toward the majority. It does not, however, demonstrate a cascade. As the preregistered exploratory structural check on over-convergence, the final-state Herfindahl concentration index (the summed squared shares of distinct final belief-sets within a group) was far higher for simulated than for human groups (chat 0.94 versus 0.62, paired difference +0.32, 95% group-bootstrap CI [+0.27, +0.37], exact sign-flip permutation $p < 0.001$, $n = 100$; reasoner 0.98 versus 0.61, +0.37 [+0.32, +0.41], $p < 0.001$, $n = 98$ groups with complete final-state records), confirming at the belief-state level that simulated groups collapse onto a single final position where human groups retain more residual disagreement.

H2b predicted that agents would deteriorate less often than humans. The same-code paired deterioration difference (simulated minus human) was +0.3 percentage points for chat (95% CI [-4.7, +4.7]) and -3.0 points for reasoner (95% CI [-7.3, +0.7]); neither interval excludes zero, so the registered direction is not established. Because a non-significant difference cannot be read as evidence of equivalence, we ran two one-sided equivalence tests (TOST) on the paired differences against a base-rate-anchored margin of +/-5 percentage points, the smallest deterioration change that would roughly double the 4.0% human rate. Within this margin, the chat difference is statistically equivalent to zero (both one-sided $p < 0.05$, $p = 0.007$ and $p = 0.034$). The reasoner difference is non-inferior in the harmful direction (one-sided $p < 0.001$), with no evidence that simulated deterioration exceeds the human rate by the prespecified margin. Symmetric equivalence fails for reasoner only because its agents deteriorate somewhat less; the point-null Bayes factors likewise favour no increase ($BF_{01} = 8.95$ for chat, 3.29 for reasoner). At a stricter +/-3-point margin the comparison is underpowered given the 4% base rate and $n = 100$ groups and is inconclusive for both modes. We therefore report H2b as no credible evidence that belief-anchored agent groups deteriorate more often than the humans they replay, and we do not interpret the reasoner's numerically lower rate as a reduction. Against the native DeliData deterioration constant of 17.6% agents would appear to deteriorate less often, but that constant reflects a different metric than the same-code 4.0% rate and is not a like-for-like comparison.

## Exploratory subjective-regime contrast (Study 2)

To test whether the over-convergence is regime-specific, we applied the same replay machinery to a subjective deliberation with no correct answer, a Japanese classroom discussion of the ethics of AI-assisted hiring (five groups, $N = 20$). This cohort is small and single-seeded and is reported as a strictly non-inferential descriptive contrast, not a confirmatory co-headline. The replay shows a convergence mismatch in the opposite direction from Study 1. On the objective Wason task, the belief-anchored replay over-converges. Here, with no correct answer to collapse onto, the same replay under-converges and holds the injected opinions more stubbornly than people do. It also closes far less of the baseline experience-based opinion separation than the human groups. The observed direction of the simulation error therefore differs between these two tasks, the kind of task-dependent distortion that a signature-level validation is meant to surface. Detailed statistics, including the experience-assortativity decay values, the label-permutation nulls, the per-arm undershoot, and an unstable memory-arm reversal, are given in Supplementary Note 1.

## Discussion

This study provides validity evidence against interpreting belief-anchored LLM groups as reproducing the distributional structure of human deliberation: in this Wason setting, they over-converge. Here, validity evidence means evidence bearing on an intended interpretation, not a claim that the system is a psychometric scale or that this constitutes a complete construct-validation programme. More narrowly, we identify estimator bias relative to human group outcomes under explicit scoring and participation definitions. Consensus does not reference correctness, so the gap is not an accuracy artifact. Although non-participation and carry-forward scoring enlarge the raw gap, participation-matched lurker-free gaps remain +34.1 points for chat and +44.4 for reasoning. The effect survives fixed-horizon stopping controls, removal of the memorizable classic answer through isomorphic reparameterization, a model-family change (Qwen3-14B), truncation checks, and seed variation. Yet near-total isomorphic agreement is predominantly wrong, so simulated consensus does not reliably indicate collective accuracy. The protocol-validity, collective-cognition, and wrong-consensus points are facets of this single failure.

For users treating language-model groups as human proxies, the bias is directional: simulated groups reach identical final answers more often even in the participation-matched, lurker-free comparison. Reading consensus, opinion spread, or minority survival from them therefore over-states agreement and under-states remaining disagreement. This structural change in the outcome distribution is not zero-mean noise removable by averaging. The distortion also appears task-sensitive: the same replay under-converges on a subjective task without a correct answer (Study 2). The differing sign across these two regimes motivates, but does not establish, task-regime dependence. This is a construct and protocol validity limit on these collective-cognition signatures, not a general claim about how agents treat dissent or represent people.

Why does the reasoning mode over-converge more than the chat mode within the same served build? We report this as a descriptive contrast, not a capability gradient. One reading, offered as a hypothesis not a demonstrated cause, is that the reasoning endpoint tracks the task's logical structure more consistently and treats the injected belief as a revisable premise rather than a sticky commitment; the transcripts are consistent with this, as reasoner agents retain their injected answer far less often than chat agents, most of their belief changes move toward the logically correct set, and their groups reach shared live beliefs in earlier rounds (Supplementary Fig. S1). But the isomorphic result shows the limit of a task-coherence reading: with the memorizable answer removed the reasoning mode still converges, yet predominantly onto wrong card sets, so its classic-task accuracy is consistent with imported competence leaking past the belief anchor rather than a validated tendency to reason its way to the truth. That agents can drift toward their model's built-in positions rather than an assigned persona is documented elsewhere (Taubenfeld et al., 2024). The classic-task peer-hidden control (Results) showed mode-dependent sensitivity: hiding peers collapsed chat consensus but barely reduced reasoning consensus, which remained high without interaction. Because peer-hidden reasoning agreement may reflect task-specific competence and the control was not run on the isomorphic task, it neither establishes a general interaction mechanism nor explains the broader mismatch. Together, the controls show that chat consensus depends strongly on peer visibility. Reasoning consensus remains high independently on the classic task, consistent with familiar-solution recovery, but high and predominantly wrong on the isomorphic task. Greater reasoning concordance therefore need not imply greater accuracy, and neither control supports interpreting greater agreement as greater competence.

Prompt-level scaffolding bounds the effect only so far. A role-fidelity instruction reduces over-convergence substantially relative to no scaffold, a shift large enough to drive a significant Page trend, but adding per-turn belief re-injection does not reduce it further and slightly reverses it. The usable lever is the instruction to hold a role, not a reminder of the initial belief on every turn.

Two recent lines of work frame the contribution. DEBATE (Chuang et al., 2025) instantiates digital-twin agents from real participants and replays group compositions on subjective questions, where a consensus cannot be decomposed into correct versus wrong; Qian et al. (2025) match agents to real participants on a collective task but do not simulate the dialogue itself, naming the distribution shift from simulated transcripts as an open problem (Supplementary Table S1). We simulate the dialogue and locate that shift in the outcome distribution. Objective ground truth then allows us to show that reasoning-mode consensus on the isomorphic task lands on wrong sets. This also distinguishes the present criterion from work characterizing agent collaboration itself: whereas Zhang et al. (2024) describe collaboration dynamics among LLM agents and Liang et al. (2024) intervene against premature convergence inside agent debate, our criterion is external, whether the resulting distribution of group outcomes matches that of the human groups being replayed. The over-convergence should not be read simply as a group-level restatement of single-model homogenization or sycophancy (Sharma et al., 2023): for the reasoning mode most belief changes move toward the logically correct card set rather than the majority, though the isomorphic result cautions against reading even this as pure task competence. Convergence is prized in multi-agent debate as a route to factual accuracy (Du et al., 2024), yet agreement among agents does not guarantee a better answer (Smit et al., 2024); the present result shows that the same convergence, when the aim is to reproduce human deliberation rather than to answer correctly, makes belief-anchored agents poor proxies for the consensus structure of a human group.

Group-decision research shows what endpoint consensus omits. Deliberation reflects informational and normative influence (Deutsch & Gerard, 1955) and biased sampling of shared over unique knowledge (Stasser & Titus, 1985; Lu et al., 2012). Majority and consistent-minority influence (Asch, 1956; Moscovici, Lage & Naffrechoux, 1969) can leave dissent or polarization rather than accuracy (Sunstein, 2002; Kerr & Tindale, 2004). Dissent is not noise: it stimulates divergent thought (Nemeth, 1986) and, with genuine participation, improves hidden-profile decisions (Schulz-Hardt et al., 2006; De Dreu & West, 2001). Accuracy also depends on interpersonal weighting (Bahrami et al., 2010), influence networks (Becker, Brackbill & Centola, 2017), and exchange structure (Navajas et al., 2018). Near-total endpoint agreement need not reproduce or certify these dynamics; minority survival, influence routes, and unshared-information sampling cannot be read from simulated transcripts.

The result is bounded in four ways. First, estimand and analytic status: the preregistered confirmatory H2a test is the all-groups paired comparison, whereas the headline participation-matched value rests on the 45 lurker-free groups, a restriction adopted after unblinding and on a subset whose groups are smaller on average (Supplementary Table S3); participation matching also does not equalize the endpoint itself, since human states are carried forward or submitted while agents are force-elicited, so a residual measurement asymmetry remains. Two considerations bound this concern rather than removing it: the independent submit-based route ($n = 98$) lands within 0.5 points of the lurker-free estimate, and the preregistered all-groups gap is itself large and significant. Second, generalizability: the confirmatory evidence rests on one objective reasoning task, the Wason selection task (Evans, 2003; Shynkaruk & Thompson, 2006), so whether the distortion transfers to tasks with different reasoning demands or demonstrability is open, and it is addressed only by the small exploratory Study 2 ($N = 20$, five groups), which shows the mirror failure of under-convergence and is a strictly non-inferential transfer signal. Third, model-and-protocol dependence: the two inference modes returned the same served model identifier and system fingerprint, so they are two inference settings of one build whose identical weights and architecture we cannot verify for a hosted model; the Qwen3-14B result is a single cross-family point on the same groups rather than a model-family matrix; the deposited DeepSeek transcripts can be re-analyzed, but exact regeneration is not guaranteed because the hosted build can change, whereas the local

Qwen3-14B runs can be rerun from the deposited configuration; and we do not read the chat-versus-reasoning difference as a capability gradient, since capability is confounded with decoding and serving. Fourth, measurement-and-exploratory limits: the primary team-performance metric is exact-set match, and although the headline consensus result does not depend on it the improvement baseline does; simulated agents almost never lurk even though the protocol permits it, so near-full participation is itself a reproduced failure and a candidate contributor to the higher consensus; and the belief anchor does not hold completely, since reasoner agents' individual accuracy rises from 11% to 85% through discussion, so some task competence leaks past the injected persona.

These limits point to a usable set of validity evidence, which we state as a checklist for agent-group studies of human deliberation. For psychologists who report or review agent-group outputs as evidence about human deliberation, the following six checks identify the evidence needed to judge whether simulated group outcomes support inferences about human collective cognition; each would have surfaced the mismatch reported here: (1) score humans and agents on a like-for-like baseline that matches participation and accounts for the difference between carried-forward and force-elicited final states, since identical code is not identical measurement; (2) use a preregistered held-out replay so that signature agreement cannot be curve-fit; (3) split any consensus into correct versus wrong against ground truth, because high agreement can coincide with high error; (4) report participation and lurking, since near-full agent participation is itself a distortion; (5) disclose the served model, version, prompt, and scaffold, given how much the endpoint depends on them; and (6) replicate across tasks of differing objectivity, to test whether the direction of the distortion differs across regimes, as the exploratory subjective-task result suggests.

These findings open as many questions as they close. First, which task properties set the direction and magnitude of the distortion: the exploratory subjective-task contrast motivates, without establishing, a systematic test that crosses objectivity, demonstrability, and answer familiarity, since a task's position on the intellective-to-judgmental continuum (Steiner, 1972; Laughlin et al., 2006) may govern whether belief-anchored groups over- or under-converge. Second, which protocol features generate or attenuate the over-convergence under a participation-matched comparison: voluntary silence, private versus public final answers, peer visibility, turn structure, and belief-anchor persistence are each manipulable, and the peer-hidden and scaffold contrasts reported here suggest they matter without isolating a mechanism. Third, what endpoint signature is sufficient to certify an agent group as a proxy for human collective cognition: full consensus alone is not, and whether jointly matching minority survival, participation, and correct-versus-wrong consensus out of sample suffices is an open empirical question. Answering these would shift the emphasis from whether agent groups agree to whether they disagree as people do.

These results carry an ethical implication for how agent groups are used as evidence. Because belief-anchored agent groups over-produce endpoint agreement, presenting their output as a stand-in for a deliberating public risks underrepresenting minority positions and overstating social agreement, so unvalidated agent consensus should not be substituted for human evidence about how much a group would actually agree. The practical implication is plain: in this deliberation setting, belief-anchored persona simulation over-produces endpoint agreement and is not yet a safe proxy for the consensus structure of human deliberation.

# Methods

## Data

We analyzed DeliData (Karadzhov, Stafford and Vlachos, 2023), a public corpus of 500 group discussions of the Wason card selection task, comprising 1,974 participants and 14,003 utterances. Each participant records a pre-discussion answer, the group discusses in free-form text chat, and each records a post-discussion answer. For every group the corpus stores a per-message solution tracker that approximates each participant's current selection over the discussion. The logically correct selection is the set containing every visible vowel card and every visible odd-number card, so team performance is scored against a ground truth rather than against opinion. The corpus uses the abstract descriptive form of the rule, not the thematic or social-contract framing on which reasoning performance differs (Cosmides & Tooby, 1992; Gigerenzer & Hug, 1992). All confirmatory analyses use this corpus and involve no new human-subjects data collection. A second cohort reuses the same replay harness on a classroom deliberation about the ethics of AI-assisted hiring (N = 20, five groups, discussion in Japanese); it is reported as an exploratory subjective-regime contrast and carries none of the confirmatory claims, because a subjective task has no correct answer against which improvement or the correctness of a consensus can be scored.

## Constrained-belief replay

A contamination probe run before the main study motivated the core design decision. Current models solve the Wason task far above the human individual baseline even under isomorphic surface reparameterization of the cards, so an agent allowed to free-solve would import a competence its assigned participant did not have, making the group accuracy gain a trivial consequence of that competence rather than a product of the deliberation. The same probe showed that an agent instantiated with a wrong pre-discussion answer retains it far more reliably under an explicit role-fidelity instruction than without one. We therefore do not let agents free-solve. Each real group is replayed by instantiating one agent per participant from that participant's pre-discussion selection, running a discussion, and eliciting each agent's final selection. An agent receives its own alias, its injected pre-discussion selection, the four visible cards, and the rule; it is never given the correct selection, the other participants' selections, or any post-discussion text, so the only information supplied beyond each agent's own injected belief is what arises within the replayed exchange. The role of peer visibility was examined by comparing peer-visible exchange with peer-hidden re-solving in a control on the classic task (Results).

## Scaffold arms and turn protocol

The role-fidelity scaffold is manipulated across three ordered arms: none, a fidelity instruction that directs the agent to reason from and defend its assigned pre-discussion selection, and a fidelity instruction combined with per-turn belief re-injection that restates the assigned selection at the head of every turn. On each turn an agent may contribute a short message or pass, so that non-participation is reproduced rather than forced, and a silent participant in a real group can be matched by an agent that passes. When an agent contributes it states its current selection, tracked to yield a per-round belief trajectory. A discussion terminates on settled belief consensus across two consecutive rounds, on an all-pass round, or on a per-group message cap. The all-pass termination and the minimum-rounds floor extend the preregistered stopping rule, which named only settled consensus and the message cap; accordingly the non-consensus-stop robustness floor reported in the Results is computed over cells that stopped by any route other than consensus, keeping it conservative. Turn order is rotated by an order seed so that no agent holds a fixed positional advantage across replays.

## Calibration and the frozen held-out protocol

Before any simulation, the corpus was split into 400 calibration groups and 100 held-out groups by a fixed hash of the group identifier, with no random-number state involved. The study was preregistered on the Open Science Framework on 4 July 2026 (https://osf.io/5jp7s), before the held-out cells were generated. Both frozen identifier lists were committed by SHA256 in that preregistration so that neither could be altered after results were seen. Calibration was permitted to tune exactly one target, the group improvement rate, and only over the scaffold arm and the temperature. The same-code human improvement rate on the 400 calibration groups was 35.8%, and calibration selected the configuration whose simulated rate came closest to it. The message cap and the minimum-rounds floor were set from the human message-count distribution rather than fitted to any signature. No other signature was inspected during calibration, and all confirmatory analyses were computed on the 100 held-out groups only, after the configuration was frozen. This single-metric, disjoint-set discipline is the load-bearing integrity control against the objection that the signature agreement was curve-fit. The frozen configuration selected the fidelity-plus-memory arm as primary, temperature 1.0, and a message cap of 25, which matches the human median. Configuration provenance, the source file and its hash config_sha256 0b27d8e9, was recorded and verified by an integrity audit (Supplementary Methods).

## Model matrix and version pinning

Two API aliases, deepseek-chat and deepseek-reasoner, were run on all 100 held-out groups with three seeds per group and arm (100 groups × 3 arms × 3 seeds × 2 inference modes = 1,800 cells), the seed varied through the prompt and agent labels rather than generator state. Both aliases returned the same served model identifier, deepseek-v4-flash, and the same system fingerprint on every successful cell, so we treat them as two inference settings of one pinned served build; the matching identifier and fingerprint document the same build across the run but do not independently verify identical weights or architecture for a hosted model. A second open-weights family, Qwen3-14B served locally with an 8,192-token context window, was run as a cross-family robustness check. Seven deepseek-reasoner cells failed on network transport errors (0.4%) and were excluded rather than backfilled, so every model-by-arm condition still retained 100 groups, while deepseek-chat had none; reasoning-mode calls used a large token budget to reduce truncation, and empty or unparseable final answers were re-elicited once and then flagged as failures. Per-condition failure, truncation, and fingerprint records are reported in Supplementary Methods.

## Signatures and statistical tests

The estimand throughout is a group-level rate in the population of DeliData Wason groups: for each signature, the proportion of groups showing that outcome, estimated from the human groups and, separately, from the simulated replays of those same groups. When we describe simulated groups as biased estimators, we mean that the simulated rate differs systematically from the human rate for the same groups under a stated scoring definition; we do not model a superpopulation beyond this corpus, and the bias is defined relative to the scoring rule in force. The confirmatory signature panel, computed by the same code for real and simulated groups and clustered on the group, comprises the group improvement rate (S1), the deterioration rate (S2), the full-consensus rate (S3), and the individual accuracy shift (S4). Team performance per group is the mean over members of an exact-set-match indicator against the correct selection, so a member counts as correct only when its final selection is identical to the ground-truth set. Full consensus is defined as identical final selections and never references the correct answer, keeping S3 independent of any accuracy metric. The available seeds per group and arm are averaged within each group, so the unit of analysis is the group (n = 100), not the cell (n = 300); this is three seeds except in the groups affected by the seven failed reasoner cells, which contribute two, and in the isomorphic replay, where seed coverage is uneven (88 groups with one seed, one with two, and 11 with three), so isomorphic per-group estimates are correspondingly less precise. Confidence intervals are group-clustered bootstraps with 10,000 resamples drawn at the group level with a fixed seed (20260724); exact sign-flip permutation tests enumerate the full sign assignment distribution by dynamic programming rather than sampling, so their p-values are exact. The confirmatory tests of H1, H2a, H2b, and H3 are one-tailed in the preregistered directions at $\alpha = 0.05$; all reported 95% confidence intervals are two-sided and are used primarily to assess effect magnitude. Permutation and bootstrap procedures are used because the group-level signature distributions are not assumed normal, so no parametric degrees of freedom are reported. The correct-versus-wrong consensus split is computed only on groups that reached full consensus, where the single shared final set is unambiguous, so it is deterministic and does not use the modal-belief tie-break. H1, the incidence of positive group accuracy change, is a one-sample centered clustered bootstrap testing whether the simulated improvement rate exceeds 50%. H2a (over-convergence) and H2b (deterioration) are paired per-group permutation tests of the simulated-minus-real difference. H3, the scaffold effect, is a trend test on the per-group aggregate signature distance across the ordered arms, the L1 distance between the simulated and human vectors over S1–S4 with each component normalized to the unit interval. The preregistration named a Jonckheere-Terpstra test; because the three arms are repeated measures on the same groups, we used the repeated-measures Page trend test instead, both monotonic-trend tests, and the H3 conclusion is unchanged. A family-wise Holm correction is applied across H1, H2a, and H2b within each inference mode, and H3 is reported separately. A signature is declared reproduced when the human value lies within the 95% bootstrap interval of the simulated estimate, deviating otherwise, a preregistered compatibility rule fixed in advance, not an equivalence test. Two planned robustness reads accompany the panel: S3 recomputed on the groups that did not stop on consensus, a conservative selected-route floor that the fixed-horizon replay then tests as a direct counterfactual, and S1 recomputed under a graded Jaccard accuracy, a check that the improvement-incidence direction does not depend on the exact-set metric. As an exploratory structural analysis, we also test whether simulated groups over-converge onto a single final belief, measured by the final-state Herfindahl index and tested paired per group by an exact sign-flip permutation, with belief states containing any token outside the group's valid cards marked invalid and excluded rather than coerced.

## Human consensus measurement and participation

In DeliData the human final state is a carry-forward tracker holding each participant's last recorded selection, so a disengaged participant keeps a possibly stale state, whereas agents are force-elicited for a final selection; scoring both with the same code therefore does not impose the same measurement. To reduce these measurement and participation asymmetries we report human full consensus under three definitions besides the carry-forward measure: active-only, restricted to participants who posted at least one message; submit-based, using each participant's actual submitted answer; and a lurker-free subset restricted to the groups in which every participant posted, on which the paired agent-versus-human risk difference is our primary effect (values in Results and Supplementary Table S2).

## Robustness replays: isomorphic task, fixed horizon, and no-peer control

Three additional held-out replays were run on the same 100 groups in the primary arm (fidelity_memory), two inference modes, using the same replay engine and final elicitation as the main run. The isomorphic replay implements the preregistered Section 4.5 contamination-robustness analysis: for each group the four Wason cards are remapped role for role to four neutral tokens (maple, birch, lantern, candle) with the logical structure held fixed, so the canonical letter-and-number answer is not available while the deductive task is unchanged. Empty or out-of-vocabulary final states fall back to the agent's last stated belief as in the main run rather than being coerced, and each replay retains all 100 groups per inference mode (truncation was negligible: at most three agent turns across the isomorphic run). The fixed-horizon replay disables the consensus and all-pass early stopping rules and runs every group to the per-group message cap, then applies the identical final elicitation, so the full-consensus rate cannot be a consequence of having stopped on convergence. The peer-hidden control used the same fixed message horizon as the peer-visible condition but withheld every message generated by other agents from each agent. All other replay settings were held constant. Each agent therefore re-solved the task without peer messages, and the endpoint was analyzed as a full-consensus rate under peer-hidden re-solving. The control was run on the classic task in both inference modes. The isomorphic and fixed-horizon replays are reported as robustness analyses and the peer-hidden run as a diagnostic control, none a new confirmatory test. All figures from them in the Results are drawn from single deduplicated manifests (first occurrence per group-seed-model key, per-group mean over seeds, unweighted mean over groups, no exclusions).

## Confirmatory status and preregistration deviations

The confirmatory analyses use the 100 held-out groups scored by the same code; the Qwen3-14B run, the isomorphic and fixed-horizon replays, and the process, mechanism, and correct-versus-wrong consensus analyses are robustness or exploratory, none a preregistered confirmatory test. Five registered-plan deviations (the same-code paired comparison in place of the native constants; the repeated-measures Page trend test in place of the registered Jonckheere-Terpstra test; the added all-pass and minimum-rounds termination, now bounded directly by the fixed-horizon replay; the execution of the preregistered Section 4.5 isomorphic analysis in this revision; and the post-unblinding adoption of the lurker-free participation-matched restriction as the headline sensitivity estimate) are disclosed in full, together with the adversarial verification, in Supplementary Methods.

## Declarations

### Ethics statement

Study 1 uses a public de-identified corpus and involves no new human-subjects data collection, and therefore required no ethics approval. For Study 2, the classroom response data were collected under the review and approval of the Waseda University Office of Research Ethics. Documentation of the review is available from the corresponding author on request. Written informed consent for use and publication was obtained from all participants, and the study followed the Declaration of Helsinki. The secondary computational analysis reported here uses only de-identified response profiles from which names and student identifiers have been removed so that no individual can be identified; free-text material is processed with a local open-weights model so that it does not leave the analysis machine.

### Declaration on the use of generative AI

During the preparation of this work the author used Claude (Anthropic) solely to polish the English language of the manuscript. All data analysis and results are the author's own genuine work and were not generated by AI. After using this tool, the author reviewed and edited the content as needed and takes full responsibility for the content and the results of the publication.

### Funding

This research was supported by the Japan Society for the Promotion of Science (JSPS KAKENHI Grant-in-Aid for Scientific Research (A), Grant Number B1K401870601; principal investigator: Masayuki Goto).

### Data availability

DeliData (Karadzhov, Stafford and Vlachos, 2023; https://doi.org/10.1145/3610056) is publicly available from its original source (https://delibot.xyz) under its own license, and we redistribute none of it. The simulated transcripts with per-cell scored outputs and served-build fingerprints, the SHA256-committed frozen calibration and held-out identifier lists, the model and scaffold configurations and prompts, and the preregistration are openly archived in a public Zenodo repository at https://doi.org/10.5281/zenodo.21318346 under CC BY 4.0. Exact regeneration of the hosted-model (DeepSeek API) transcripts is not guaranteed because the served build can change, so we deposit the recorded transcripts and fingerprints rather than a rerun script for those. The Study 2 per-response classroom data are withheld for participant confidentiality and are available from the corresponding author under a data-use agreement.

## Code availability

The analysis and replay code is openly archived in the same public Zenodo repository (https://doi.org/10.5281/zenodo.21318346) under the MIT License, and the local Qwen3-14B runs are reproducible from the deposited configuration.

## Competing interests

The author declares no competing interests.

## Author contributions and acknowledgements

T.S.: Conceptualization, Methodology, Software, Formal analysis, Investigation, Data curation, Writing – original draft, Writing – review & editing. John Maurice Gayed and Hidehiro Kanemitsu contributed to investigation and to reviewing and editing the manuscript, and Masayuki Goto supervised the project, acquired funding, and reviewed and edited the manuscript. The peer-reviewed journal version of this work is co-authored with them.

## Table 1

**Table 1. Hypothesis tests on the 100 held-out groups.** Estimates are on the primary preregistered arm (fidelity_memory) and are group-clustered; the confirmatory directional p-values are Holm-corrected across the family {H1, H2a, H2b}, and H3 is a separate trend test. For H2b, which returned no significant directional difference, we report the paired difference with its 95% CI and a two-one-sided-tests (TOST) equivalence result; these TOST p-values are reported separately and are not part of the Holm family.

| Hypothesis | deepseek-chat | deepseek-reasoner | Verdict |
|---|---|---|---|
| **H1: incidence of positive accuracy change > 50%** | obs − 0.5 = +0.107, p_holm = 0.0029 | +0.353, p_holm < 0.001 | supported both (threshold) |
| **H2a: over-converge vs human (paired)** | lurker-free +34.1 pts [19.3, 48.9]; all-groups +62.0 [53.0, 70.7], p_holm < 0.001, h = 1.35 | lurker-free +44.4 pts [29.6, 59.3]; all-groups +72.0 [63.3, 80.3], p_holm < 0.001, h = 1.71 | supported both (headline result) |
| **H2b: deterioration not greater than human (paired)** | +0.3 pts [−4.7, +4.7]; equivalent within ±5 pts (TOST p = 0.007 / 0.034) | −3.0 pts [−7.3, +0.7]; non-inferior at ±5 pts (one-sided p < 0.001) | no credible evidence of greater deterioration |

| **H3: scaffold trend (Page L)** | $L = 1267.5$, $p < 0.001$ | $L = 1250.5$, $p < 0.001$ | trend yes, monotonic order no |
|---|---|---|---|

The across-arm signature rates (S1 improve, S2 worse, S3 full consensus) are given in Supplementary Table S4, and the protocol robustness replays with the peer-hidden diagnostic control in Supplementary Table S5. Human full consensus under alternative measurement and participation definitions is given in Table 2; estimate confidence intervals and the remaining robustness and sensitivity checks are given in Supplementary Table S2.

## Table 2

**Table 2. Human full consensus depends on how participation and final states are scored.** Full-consensus rate among the same 100 held-out DeliData groups under four scoring definitions, computed by the same code used for the simulated groups. The rate varies by more than a factor of two across definitions, so the operationalization of consensus is part of the validity argument rather than a reporting detail. The lurker-free row is the participation-matched basis for the headline paired comparison; it is a post-unblinding sensitivity estimate, not a preregistered confirmatory quantity.

| Definition | Full consensus (%) | Eligible n | Analytical role |
|---|---|---|---|
| **Carry-forward (corpus solution tracker)** | 24.0 | 100 | corpus default; human row of Supplementary Table S4 |
| **Submit-based (actual submitted answers)** | 52.0 | 98 | measurement sensitivity |
| **Active-only (posted at least one message)** | 57.0 | 100 | participation sensitivity |
| **Lurker-free groups (every member posted)** | 51.1 | 45 | primary participation-matched paired comparison |

## Supplementary information

### Supplementary Methods

**Model matrix: per-condition failure, truncation, and fingerprint records.** Both requested API aliases (deepseek-chat, deepseek-reasoner) returned the same served model identifier, deepseek-v4-flash, and the same system fingerprint (fp_8b330d02d0_prod0820_fp8_kvcache_20260402) on every successful cell (900 chat, 893 reasoner), so we treat them as two inference settings of one recorded served build; the matching identifier and fingerprint document a consistent build across the run but do not pin the model in the sense of an independently retrievable checkpoint, and do not verify identical weights or architecture for a hosted model. Seven deepseek-reasoner cells failed on network transport errors (0.4%, six IncompleteRead and one RemoteDisconnected) and were excluded rather than backfilled, so every model-by-arm condition still retained 100 groups, while deepseek-chat had none. Length-finish truncations affected 223 of 13,770 reasoner calls (1.6%), none for chat, and failure and truncation rates are reported per condition; removing the truncated reasoner cells raises the reasoner full-consensus rate from 96.0% to 97.5%, so truncation deflates convergence rather than manufacturing it.

**Confirmatory status and preregistration deviations.** The confirmatory analyses use the 100 held-out groups scored by the same code; the Qwen3-14B run, the isomorphic and fixed-horizon replays, and the process, mechanism, and correct-versus-wrong consensus analyses are robustness or exploratory, none a preregistered confirmatory test.

Five points about the registered plan are disclosed. First, the H2 and H3 tests use the same-code paired comparison rather than the native constants, a decision fixed in the analysis code on 7 July 2026 before the held-out cells were unblinded on 8 July. Second, H3 uses the repeated-measures Page trend test rather than the registered Jonckheere-Terpstra test, both monotonic-trend tests. Third, the stopping rule adds all-pass and minimum-rounds termination beyond the registered consensus-and-cap rule, which we now bound directly with the fixed-horizon replay rather than only with the selected-route floor. Fourth, the preregistered Section 4.5 isomorphic-reparameterization analysis was not part of the original confirmatory run and is executed in this revision as a robustness replay. Fifth, the preregistered confirmatory H2a test is the all-groups paired comparison, and it is the quantity carrying the Holm-corrected p-values; the lurker-free participation-matched restriction that the Results treat as the headline was adopted during revision, after the held-out results were unblinded, and its intervals are therefore reported as a participation-sensitivity estimate rather than a preregistered confirmatory quantity.

**Adversarial verification.** We verified the held-out results by re-deriving every reported quantity from the raw per-cell score fields, auditing the protocol against the frozen configuration, and stress-testing the headline under three adversarial robustness attacks. The integrity audit confirmed that the calibration and held-out sets were disjoint, that both identifier-list hashes matched the values committed in the preregistration, that no held-out group leaked into calibration, and that failed cells were counted and not backfilled. The three attacks targeted stopping-rule circularity, metric and baseline choice, and sensitivity to truncation, exclusion, and seed. Every reported number reproduced from the raw fields, and the headline result held under all three attacks. These checks are deterministic and reproducible from the deposited analysis code. The same-code paired test, adopted as primary because the preregistration was internally inconsistent between the same-code principle and named native constants, was fixed in the analysis code before the held-out cells were unblinded, so it preceded any sight of the confirmatory outcomes and could not have been chosen to favor a result.

## Supplementary Table S1

**Supplementary Table S1. This study relative to two recent digital-twin deliberation benchmarks.**

| Feature | DEBATE (2025) | Qian et al. (2025) | This study |
|---|---|---|---|
| **Agents from real participants** | yes | yes | yes |
| **Group dialogue simulated** | yes | no | yes |
| **Objective ground truth** | no (subjective) | partial | yes (Wason) |
| **Correct vs wrong consensus separable** | no | no | yes |
| **Preregistered held-out test** | no | no | yes |

## Supplementary Table S2

**Supplementary Table S2. Primary-arm estimates with 95% group-clustered bootstrap confidence intervals (a) and remaining robustness and sensitivity checks (b).** Fixed-horizon, isomorphic, and no-peer full-consensus rates appear in Supplementary Table S5 and Fig. 3. Cross-family: Qwen3-14B S3 65.7% (paired +41.7 points against the carry-forward human rate, $p < 0.001$ nominal, Cohen's $h = 0.87$).

(a) Estimates and 95% CIs (primary arm, percent of groups):

| Quantity | deepseek-chat | deepseek-reasoner | Human (same-code) |
|---|---|---|---|
| **S1 improve** | 60.7 [53.7, 68.0] | 85.3 [80.3, 90.0] | 41.0 |
| **S3 full consensus** | 86.0 [81.7, 90.0] | 96.0 [93.7, 98.0] | 24.0 (carry-forward) |
| **Participation-matched paired over-convergence (45 lurker-free groups)** | +34.1 [19.3, 48.9] | +44.4 [29.6, 59.3] | n/a |
| **Submit-based paired over-convergence, groups with at least two human submitters (n = 98)** | +34.0 [24.1, 44.2] | +43.9 [34.0, 53.7] | n/a |

The submit-based row compares each group's agent full consensus against the human submit-based full-consensus rate of 52.0%; human consensus is computed among submitters whereas agent consensus includes all assigned agents, so this sensitivity analysis reduces the endpoint-measurement asymmetry without fully equalizing participation or endpoint membership. It is a post-unblinding sensitivity estimate, not a preregistered confirmatory quantity.

(b) Robustness and sensitivity (primary arm):

| Check | deepseek-chat | deepseek-reasoner |
|---|---|---|
| **Isomorphic correct / wrong consensus (of all groups)** | 36.7 / 43.3 | 24.7 / 74.0 |
| **Non-consensus-stop S3 floor** | 75.0 | 94.3 |
| **Anchor retention (kept injected belief, %)** | 34.6 | 17.1 |
| **Graded-Jaccard S1 (human 77.0)** | 86.0 | 94.7 |
| **Per-seed S3 spread** | at most 3 points | at most 3 points |
| **S4 individual accuracy shift (human 11% to 33%)** | 11% to 58% | 11% to 85% |
| **H3 scaffold distance, closest to farthest: fidelity, fidelity_memory, none** | 0.998, 1.073, 1.650 | 1.604, 1.709, 2.059 |

The H3 distances are L1 distances to the human S1–S4 signature vector with each component normalized to the unit interval, so plain fidelity sits closest to human behaviour in both modes.

## Supplementary Table S3

**Supplementary Table S3. Composition of the participation-matched (lurker-free) subset relative to the full held-out sample.** Pre-outcome descriptors only; comparisons are descriptive and no inferential tests were performed. The all-held-out column contains the 45 lurker-free groups, so the has-lurker complement is reported as the independent contrast.

| Descriptor | Lurker-free, n = 45 | Has-lurker complement, n = 55 | All held-out, n = 100 |
|---|---|---|---|
| **Group size, mean (SD), range** | 3.20 (0.94), 2 to 5 | 4.40 (0.89), 3 to 7 | 3.86 (1.09), 2 to 7 |
| **Distinct initial answers, mean (SD)** | 2.73 (0.81) | 3.35 (0.75) | 3.07 (0.83) |
| **Mean within-group correct-starter fraction** | 9.0% | 13.0% | 11.2% |
| **Groups with at least one correct starter** | 26.7% | 38.2% | 33.0% |
| **Initially unanimous groups** | 1/45 (2.2%) | 0/55 (0.0%) | 1/100 (1.0%) |

The 45 lurker-free groups contributing to the participation-matched estimate form a smaller-group subset of the 100 held-out groups. Their mean group size was 3.20 (SD 0.94, range 2 to 5), compared with 3.86 (SD 1.09, range 2 to 7) across all held-out groups. This pattern is consistent with conditioning on every member posting, but the descriptive comparison does not identify the cause of selection. Only one of the 45 lurker-free groups began in full agreement, so 44 began with more than one distinct pre-discussion answer. The mean within-group proportion of correct starters was 9.0%, and 26.7% of groups contained at least one correct starter, compared with 11.2% and 33.0% respectively across all held-out groups. Pre-existing unanimity and elevated initial correctness therefore do not characterize the subset. Mean initial belief diversity was also lower, at 2.73 rather than 3.07 distinct pre-discussion answers per group, alongside the smaller group size. Because agent and human outcomes are compared within the same group, group size and starting beliefs do not create a between-side compositional imbalance within the 45 pairs. Pairing does not, however, rule out effect modification: smaller group size could affect agent and human consensus differently and thereby influence the magnitude of the paired difference. The estimate should be interpreted as a participation-matched sensitivity estimate for predominantly two- to five-member groups that generally began in disagreement and with low initial accuracy, rather than as evidence that the same difference applies to larger groups. Participation matching also does not remove the remaining difference between the human and agent endpoint measurements.

## Supplementary Table S4

**Supplementary Table S4. Across-arm signature rates (S1 improve, S2 worse, S3 full consensus) on the 100 held-out groups.** Simulated rates are group-clustered means by model and scaffold arm; human rates are recomputed by the same code on the same groups, with human S3 under the corpus carry-forward measure. Percent of groups.

| Model | Arm | S1 improve | S2 worse | S3 full consensus |
|---|---|---|---|---|
| **deepseek-chat** | none | 82.3 | 4.0 | 98.7 |
| **deepseek-chat** | fidelity | 59.7 | 4.0 | 81.7 |
| **deepseek-chat** | fidelity_memory (primary) | 60.7 | 4.3 | 86.0 |
| **deepseek-reasoner** | none | 100.0 | 0.0 | 100.0 |
| **deepseek-reasoner** | fidelity | 80.7 | 2.0 | 96.0 |
| **deepseek-reasoner** | fidelity_memory (primary) | 85.3 | 1.0 | 96.0 |
| **Qwen3-14B (cross-family)** | fidelity_memory (primary) | 37.7 | 10.7 | 65.7 |
| **Human (same-code)** | n/a | 41.0 | 4.0 | 24.0 |

## Supplementary Table S5

**Supplementary Table S5. Full-consensus rate under protocol robustness replays and the peer-hidden diagnostic control (primary arm).** Fixed-horizon and isomorphic replays are robustness analyses; the no-peer / solitary run is a diagnostic control that measures agreement among agents separately re-solving the classic task, not a robustness check. Percent of groups; n = 100 per cell. These rates are shown in Fig. 3.

| Variant | deepseek-chat | deepseek-reasoner |
|---|---|---|
| **Main (raw)** | 86.0 | 96.0 |
| **Fixed-horizon (early stops disabled)** | 87.0 | 100.0 |
| **Isomorphic reparameterization** | 80.0 | 98.7 |
| **No-peer / solitary (diagnostic control)** | 31.0 | 91.0 |

## Supplementary Figure S1

`![Supplementary Figure S1](figures/si_fig_trajectory.png)`

**Supplementary Figure S1. Agent-only cumulative consensus trajectories (exploratory).** Cumulative share of the 100 held-out groups whose per-round CURRENT beliefs were all identical at least once by round t, on the primary arm, per seed and averaged within group before averaging across groups; shaded

bands are 95% group-level bootstrap intervals. By round two, 76.0% of reasoner groups had held a shared live belief against 26.7% for chat (round three: 92.3% against 42.0%; any round: 95.3% against 62.0%). Human trajectories are not overlaid because DeliData belief states update per message whereas agent states are sampled per round, so the two are not on a common time base. The curves describe when shared live beliefs were first observed within this protocol; they do not identify a social-influence mechanism and do not establish that convergence was premature relative to humans. The final-answer full-consensus rates (chat 86.0%, reasoner 96.0%) come from the separate forced final elicitation and are not a further round of these curves.

## Supplementary Note 1. Study 2 detailed statistics

Study 2 applied the same replay machinery to a subjective classroom deliberation about the ethics of AI-assisted hiring (five groups, N = 20, 8 participants with job-hunting experience and 12 without). Responses were recorded at three times, before reading (T1), after reading but before discussion (T2), and after discussion (T3), with agents instantiated from each participant's T2 profile and the post-discussion T3 rating held out as the validation target. All free text was processed with a local open-weights model. This cohort is small and single-seeded, so it is reported as an exploratory descriptive contrast, not an inferential effect.

The signature is experience-assortativity decay: the Euclidean separation D between experienced and inexperienced opinion centroids in five-item Likert space, which deliberation is expected to consume. The two groups started apart (D(T1) = 2.592) and the human groups largely closed the gap by T3 (D = 0.599), a decay of +1.993 well separated from a label-permutation null (20,000 permutations, $p = 0.0013$).

The simulation reproduced the direction of this convergence but fell far short of its magnitude. Across the three scaffold arms, the two without per-turn belief re-injection closed only about a third of the human separation: simulated decay was +0.682 (none) and +0.689 (fidelity), 34% and 35% of the human value, and 24% once both trajectories are re-anchored at T2 where the agents actually begin. Neither arm was distinguishable from the label-permutation null ($p = 0.291$ and 0.185), and the residual separation (about 1.91) remained more than three times the human post-discussion value. The undershoot is not an artifact of the fidelity instruction, because the none arm, which carries no stay-in-character instruction, undershoots essentially identically.

One exploratory observation runs counter to Study 1: per-turn belief re-injection produced the largest decay (+1.277, 64% of the human value, marginal $p = 0.064$) and came closest to the human profile, the opposite of its reining-in role on the objective task. With $N = 20$ and one stochastic draw per arm this ordering is not stable and is reported as an observation only.